\documentclass[11pt]{article}
\usepackage[a4paper, margin=1in]{geometry} 
\usepackage{graphicx}
\usepackage{twocolceurws}
\usepackage{titling}
\usepackage{authblk}
\usepackage{hyperref}
\usepackage{breqn}
\title{\textbf{A Comparative Study of Machine Unlearning Techniques for Image and Text Classification Models}}

\author{Omar M. Safa, Mahmoud M. Abdelaziz, Mustafa Eltawy, Mohamed Mamdouh, \\
Moamen Gharib, Salaheldin Eltenihy, Nagia M. Ghanem, and Mohamed M. Ismail}

\affil{
    \textit{Computer and Systems Engineering Department} \\ 
    \textit{Faculty of Engineering, Alexandria University} \\ 
    \textit{Alexandria, Egypt}
}
\date{}
\begin{document}
\maketitle

\begin{abstract}
Machine Unlearning has emerged as a critical area in artificial intelligence, addressing the need to selectively remove learned data from machine learning models in response to data privacy regulations. This paper provides a comprehensive comparative analysis of six state-of-the-art unlearning techniques applied to image and text classification tasks. We evaluate their performance, efficiency, and compliance with regulatory requirements, highlighting their strengths and limitations in practical scenarios. By systematically analyzing these methods, we aim to provide insights into their applicability, challenges, and trade-offs, fostering advancements in the field of ethical and adaptable machine learning.\end{abstract}

\section{Introduction}

The rapid expansion of machine learning (ML) applications has raised significant concerns over data privacy. With the integration of ML models across industries, regulations like the GDPR and CCPA mandate that individuals can request the removal of their personal data. This creates a challenge for organizations to comply with privacy laws while maintaining model performance. Traditional ML training often results in models memorizing sensitive data, and removing this data typically requires retraining from scratch, which is resource-intensive. Machine unlearning has been proposed as a solution to allow models to “forget” specific data efficiently without complete retraining, preserving performance and ensuring privacy compliance. Previous work on unlearning in classification models has categorized the problem into three distinct scenarios: (1) \textbf{Full-class unlearning},
which entails removing an entire class from the
model, with the forget set containing all instances
of a specific class; (2) \textbf{Sub-class unlearning},
which focuses on forgetting a subset of instances
within a single class; and (3) \textbf{Random forgetting}, which involves removing arbitrary instances across multiple classes. Despite progress in this field, a critical gap remains in the literature: the absence of a comprehensive, unified study comparing unlearning
techniques across diverse datasets and models.
Existing research typically evaluates methods on
specific datasets or tasks, leading to fragmented
insights and a lack of standardized benchmarks
for performance comparison. This paper aims to bridge this gap by conducting a comparative analysis of prominent machine unlearning techniques in the context of image and text classification tasks. By evaluating these methods across multiple datasets and offering a unified framework for comparison, we aim to provide valuable insights into the trade-offs, strengths, and weaknesses of each approach.
\section{Related Work}

Foster, Schopf, and Brintrup \cite{foster2023fast} introduced an unlearning algorithm called Selective Synaptic Dampening \textbf{SSD} that uses weight sensitivity to adjust model parameters selectively for forgetting specific data. The method identifies weights heavily influenced by the "forget set" compared to the remaining data and dampens these weights, reducing their impact on predictions. Weight sensitivities are computed using the Fisher Information Matrix (FIM), enabling targeted adjustments to the most sensitive model parameters. Graves, Nagisetty, and Ganesh \cite{graves2021amnesiac} proposed an unlearning method called \textbf{Mislabel Unlearning}, a simple approach in which the labels of data in the forget set are randomly changed to those of other classes. The modified data is then used to fine-tune the model for a few epochs, enabling selective forgetting. Chundawat, et al \cite{chundawat2023zeroshot} proposed an unlearning method called \textbf{Incompetent Teacher} that uses selective knowledge transfer between student and teacher models. The student model learns from the forget set under the influence of an incompetent teacher, while a competent teacher provides corrective guidance for the retain set. Kurmanji et al. \cite{kurmanji2023unbounded} proposed \textbf{SCRUB}, an unlearning method based on the teacher-student framework. Similar to the "Incompetent Teacher" approach, SCRUB uses a student model that learns from a teacher model. However, SCRUB modifies loss function to increase error on the forget set while optimizing accuracy on the retained data. This is achieved through alternating "max-steps" (focused on forgetting) and "min-steps" (focused on retaining), with additional steps to restore performance on the retain set. Tarun et al. \cite{tarun2023fast} introduced the \textbf{UNSIR} algorithm. This approach consists of three steps: generating an error-maximizing noise matrix, impairing the model by training it with this noise on a subset of data, and repairing the model on the retained data. The noise matrix, which maximizes error for the targeted class, is key to disrupting the model’s ability to recall the forgotten data. Following the impair step, the repair step ensures that the model retains its performance on the remaining data.


\section{Methodology}

This study evaluates machine unlearning techniques using image and text classification models, focusing on the trade-off between retaining accuracy on non-forgotten data and effectively forgetting target data. The experiments involve fine-tuning pre-trained models, applying unlearning processes, and measuring performance using key metrics.

\subsection{Models and Datasets}

\textbf{Image Classification Models:} 
\begin{itemize}
    \item \textbf{ResNet18 \cite{he2016deep}:} A lightweight residual convolutional network, known for its efficient training.
    \item \textbf{ViT(google/vit-base-patch16-224) \cite{wu2020visual}:} Vision Transformer \cite{dosovitskiy2021image}, which treats image patches as tokens and utilizes transformer architecture.
\end{itemize}

\textbf{Text Classification Model:} 
\begin{itemize}
    \item \textbf{MARBERT \cite{abdul-mageed-etal-2021-arbert}:} An Arabic variant of BERT, pre-trained on 128 GB of Arabic text data.
\end{itemize}

\textbf{Image Datasets:} 
\begin{itemize}
    \item \textbf{CIFAR-10 \cite{krizhevsky2009learning}:} Contains 60,000 color images of size 32x32 pixels across 10 classes (e.g., airplanes, cars, birds, etc.). It is used for random forgetting experiments, with 50,000 training and 10,000 test images.
    \item \textbf{CIFAR-100:} A more challenging dataset with 100 classes and 600 images per class. Used for full-class unlearning experiments, it is split into 50,000 training and 10,000 test images.
\end{itemize}

\textbf{Text Dataset:} 
\begin{itemize}
    \item \textbf{Hotel Arabic Reviews Dataset (HARD) \cite{Elnagar2018}:} Contains 93,700 Arabic hotel reviews (only 50,000 were used), labeled from 1 to 5, split into 40,000 training samples and the 10,000 for testing.
\end{itemize}
In our experiments, the models were fine-tuned on the datasets mentioned above. The experiment aimed to perform Full-Class Unlearning and/or Random Forgetting using six unlearning methods: SSD, Incompetent Teacher, SCRUB, UNSIR, and Mislabel. Retraining from scratch was also included as a baseline for comparison.

\subsection{Evaluation Metrics}

The effectiveness of machine unlearning is assessed using the following metrics:

\subsubsection{Relative Retain Accuracy ($A_r$)}
Retain accuracy measures the performance of the unlearned model on data not targeted for forgetting. It is computed as:
\begin{equation}
A_r = \frac{A_u}{A_b} \times 100,\
\end{equation}
where $A_u$ is the accuracy of the unlearned model on the retain set, and $A_b$ is the baseline accuracy of the original model. An ideal unlearning method should preserve $A_r$ close to 100\%.

\subsubsection{Relative Forget Accuracy ($A_f$)}
Forget accuracy quantifies the model's accuracy on the forget set after unlearning. Lower $A_f$ values indicate effective forgetting. It is defined as:
\begin{equation}
A_f = \frac{A_u}{A_b} \times 100,\
\end{equation}

\subsubsection{Membership Inference Attack (MIA)}
MIA \cite{shokri2017membership} evaluates the model's susceptibility to leaking information about the forget set. Based on logistic regression, MIA returns a probability score indicating the likelihood of a data sample being part of the training set. The optimal MIA value is defined as:
An optimal value for MIA might appear to be closer to 0, however as argued by Chundawat et. al \cite{chundawat2023zeroshot}, an abnormally small probability provides as much information to an
attacker as an exceedingly large probability. Thus, optimal values for MIA are somewhere
between and ideally very close to the values that would be produced by a retrained model.

\subsubsection{Zero Retrain Forgetting (ZRF)}
ZRF evaluates the randomness of predictions on the forget set without requiring a retrained model. It is computed using Jensen-Shannon divergence ($JS$) \cite{lin1991divergence} as:
\begin{dmath}
JS(M(x), T_d(x)) = 0.5 \cdot KL(M(x) || m) + \\ 0.5 \cdot KL(T_d(x) || m),
\end{dmath}
where $m = \frac{M(x) + T_d(x)}{2}$ is the mean distribution, $M(x)$ is the output of the unlearned model, and $T_d(x)$ is the output of an incompetent teacher model. The ZRF score is given by:
\begin{equation}
ZRF = 1 - \frac{1}{n_f} \sum_{i=1}^{n_f} JS(M(x_i), T_d(x_i)),
\end{equation}
where $n_f$ is the number of samples in the forget set. ZRF values closer to 1 indicate highly randomized predictions, signifying effective forgetting.

\subsubsection{Computation Time}
The time required for applying the unlearning process is recorded as a practical metric. Efficient computation time is crucial for scalability, especially in compliance with privacy laws such as GDPR and CCPA.

\section{Experiments}
In our experiments, the models were fine-tuned on
the datasets mentioned above. The experiment
aimed to perform Full-Class Unlearning and/or
Random Forgetting using six unlearning methods:
SSD, Incompetent Teacher, SCRUB, UNSIR, and
Mislabel. Retraining from scratch was also included as a baseline for comparison.
For the full-class forgetting process, the class to be forgotten for each of the ResNet18
and ViT image classification models is (Rocket).
Whereas for the MARBERT sentiment analysis classification task, the experiments are done for each of the 5 classes in HARD dataset (Rating from 1 to 5).
\subsection{Hyperparameters}
\textbf{For Full-Class Unlearning and Random Forgetting in ResNet}:\\
For Selective Synaptive Dampening the values shown correspond to $\alpha = 15$, $\gamma = 1$.\\
For The Incompetent Teacher the results correspond to epochs = 1 and learning rate = 0.1.\\
For Mislabel Unlearning the values shown correspond to epochs = 1 and learning rate = 0.0001.\\
\textbf{For Full-Class Unlearning and Random Forgetting in ViT}:\\
For Selective Synaptic Dampening the values shown correspond to $\alpha = 25$, $\gamma = 1$.\\
For The Incompetent Teacher the results correspond to epochs = 1 and learning rate = 0.0002.\\
For Mislabel Unlearning the values shown correspond to epochs = 1 and learning rate = 0.0001\\
\textbf{For Full-Class Unlearning in  MARBERT}:\\
For Selective Synaptive Dampening the values shown correspond to $\gamma = 1.45$ for all classes with the following $\alpha$ values for each class:
\begin{itemize}
    \item Rating 1: $\alpha = 14$
    \item Rating 2: $\alpha = 2$
    \item Rating 3: $\alpha = 2$
    \item Rating 4: $\alpha = 2$
    \item Rating 5: $\alpha = 6$
\end{itemize}
For The Incompetent Teacher the results correspond to epochs = 1 and two learning rates: 1e-4 and 2e-5.\\
For Mislabel Unlearning the values shown correspond to epochs = 1 and learning rate = 0.0001\\
For UNSIR, the values shown correspond to one impair-repair step, learning rate for impair = 0.0001, learning rate for repair = 0.0001 and lambda for noise generation was 0 for 4 \& 5 classes and was 0.1 for 1 \& 2 \& 3 classes.\\
For all experiments, default hyper-parameters suggested in the paper \cite{kurmanji2023unbounded} were used for SCRUB, learning rate = 0.0001, alpha=0.001, gamma=0.99, with extending number of unlearning epochs from 2 to 4.
\section{Results}
First for Image Classification Models

\begin{table*}[ht]
    \centering
    \resizebox{\textwidth}{!}{
    \begin{tabular}{ccccccc}
        \hline
        \textbf{Metric} & \textbf{\boldmath$Acc_t$} & \textbf{\boldmath$Acc_r$} & \textbf{\boldmath$Acc_f$} & \textbf{ZRF} & \textbf{MIA} & \textbf{Time} \\
        \hline
        \textbf{Baseline} & 100.00 & 100.00 & 100.00 & 0.9783 & 0.946 & - \\
        \textbf{Retrain} & 94.06 & 95.04 & \textbf{0} & 0.9955 & 0.014 & 5840 \\
        \textbf{SSD} & 98.61 & 99.63 & \textbf{0} & 0.9932 & 0.008 & 214 \\
        \textbf{Incompetent Teacher} & 96.79 & 97.83 & \textbf{0} & 0.9989 & \textbf{0} & \textbf{187} \\
        \textbf{SCRUB} & \textbf{98.97} & \textbf{100.02} & \textbf{0} & 0.9895 & 0.21 & 528 \\
        \textbf{UNSIR} & 96.62 & 97.21 & 43.99 & 0.9868 & 0.058 & 258 \\
        \textbf{Mislabel} & 96.81 & 97.84 & \textbf{0} & \textbf{0.9989} & \textbf{0} & 203 \\
        \hline
    \end{tabular}
    }
    \caption{Results for forgetting class (Rocket) from ResNet18 finetuned on Cifar100.}
    \label{tab:rocket-results}
\end{table*}

\begin{table*}[ht]
    \centering
    \resizebox{\textwidth}{!}{
    \begin{tabular}{cccccccc}
        \hline
        \textbf{Metric} & \textbf{\boldmath$Acc_t$} & \textbf{\boldmath$Acc_r$} & \textbf{\boldmath$Acc_f$} & \textbf{ZRF} & \textbf{MIA} & \textbf{Time} \\
        \hline
        \textbf{Baseline} & 100.00 & 100.00 & 100.00 & 0.9863 & 0.7850 & -\\
        \textbf{Retrain} & 105.10 & 105.52 & \textbf{0} & 0.9961 & 0.0117 & 12451 \\
        \textbf{SSD} & 98.94 & 99.96 & \textbf{0} & 0.9941 & 0.0300 & 2264 \\
        \textbf{Incompetent Teacher} & 98.00 & 98.20 & 11.49 & \textbf{0.9992} & \textbf{0} & \textbf{1270} \\
        \textbf{SCRUB} & 101.88 & 102.91 & \textbf{0} & 0.9839 & 0.3000 & 8339 \\
        \textbf{UNSIR} & 93.88 & 102.31 & \textbf{0} & 0.9890 & 0.1117 & 3778 \\
        \textbf{Mislabel} & \textbf{107.96} & \textbf{108.27} & \textbf{0} & 0.9981 & 0.0050 & 5717 \\
        \hline
    \end{tabular}
    }
    \caption{Results for forgetting class (Rocket) from ViT finetuned on Cifar100.}
    \vspace{10pt}
    \label{tab:rocket-results}
\end{table*}
\subsection{Full Class Forgetting}
\subsubsection{ResNet18 Results}
For ResNet18, SCRUB demonstrated the best performance in terms of both test accuracy and retain accuracy in the full-class forgetting scenario. The improvement in retain accuracy relative to the baseline is attributed to the inclusion of a cross-entropy loss term in SCRUB’s loss function, which enhances accuracy during the unlearning process.

Selective Synaptic Dampening (SSD) also achieved very high retain accuracy, trailing only SCRUB. SSD was only slightly behind Incompetent Teacher and Mislabel Unlearning in the ZRF and MIA metrics. Furthermore, SSD was more than twice as fast as SCRUB (although slightly slower than Mislabel Unlearning) and achieved marginally better ZRF and MIA metric values, establishing SSD as a balanced and efficient method for this experiment.

UNSIR, while achieving stronger-than-retrain results across all metrics, trailed the other methods in overall performance. Interestingly, UNSIR retained some performance on the forget set, highlighting UNSIR unability to completely forget classes while keeping constant retain accuracy.

Even the slowest algorithm applied in this experiment achieved approximately a 95\% speed-up compared to naive retraining, with better retain accuracy and nearly equivalent performance in the MIA and ZRF metrics.

\subsubsection{ViT Results}
Mislabel Unlearning demonstrated the most significant improvement in relative test and retain accuracy, with ZRF and MIA values closely aligning with those of the retrain algorithm. This underscores its effectiveness in unlearning while maintaining high accuracy. SCRUB, while achieving strong ZRF values, exhibited elevated MIA metrics, suggesting reliable forgetting but reduced security. Selective Synaptic Dampening (SSD), although unable to fully preserve overall accuracy, offered a balanced unlearning approach with low MIA. Its performance, combined with being twice as fast as Mislabel Unlearning and four times faster than SCRUB, positions SSD as a more efficient alternative. In contrast, Incompetent Teacher failed to completely unlearn the Rocket class; however, this moderate unlearning resulted in a secure model with zero MIA, enhancing its robustness.
It is important to note that several methods improved the overall accuracy of the model, with Mislabel Unlearning demonstrating the most substantial increase. This improvement can be attributed to two key factors: the baseline model not being fully trained to saturation and the ViT model's reliance on transfer learning. Consequently, additional training epochs, whether for learning or unlearning, facilitate further fine-tuning and enhance the model's accuracy.

\subsection{Random Forgetting}
\begin{table*}[h!]
    \centering
    \resizebox{\linewidth}{!}{
    \begin{tabular}{ccccccc}
        \hline
        \textbf{Metric} & \textbf{\boldmath$Acc_t$} & \textbf{\boldmath$Acc_r$} & \textbf{\boldmath$Acc_f$} & \textbf{ZRF} & \textbf{MIA} & \textbf{Time} \\
        \hline
        \textbf{Baseline} & 100 & 100 & 100 & 0.7925 & 0.7692 & - \\
        \textbf{Retrain} & \textbf{100} & \textbf{100} & 99 & 0.7959 & 0.7484 & 757 \\
        \textbf{SSD} & \textbf{100} & \textbf{100} & 100 & 0.7925 & 0.7692 & 243 \\
        \textbf{Incompetent Teacher} & 99 & 99 & 98 & 0.9613 & 0.4468 & 188 \\
        \textbf{SCRUB} & 83 & 83 & 104 & \textbf{0.9999} & 0.9900 & 326 \\
        \textbf{Mislabel} & 97 & 97 & \textbf{97} & 0.9708 & \textbf{0.4127} & \textbf{194} \\
        \hline
    \end{tabular}
    }
    \caption{Random Forgetting 0.05 of data from ResNet.}
    \label{tab:resnet_random_forgetting_with_scrub_0.05_transposed}
\end{table*}

\begin{table*}[h!]
    \centering
    \resizebox{\linewidth}{!}{
    \begin{tabular}{ccccccc}
        \hline
        \textbf{Metric} & \textbf{\boldmath$Acc_t$} & \textbf{\boldmath$Acc_r$} & \textbf{\boldmath$Acc_f$} & \textbf{ZRF} & \textbf{MIA} & \textbf{Time} \\
        \hline
        \textbf{Baseline} & 100 & 100 & 100 & 0.7934 & 0.7632 & - \\
        \textbf{Retrain} & \textbf{100} & \textbf{100} & 99 & 0.8007 & 0.7413 & 722 \\
        \textbf{SSD} & \textbf{100} & \textbf{100} & 100 & 0.7934 & 0.7632 & 246 \\
        \textbf{Incompetent Teacher} & 97 & 97 & 96 & 0.9804 & 0.4164 & \textbf{195} \\
        \textbf{SCRUB} & 82 & 82 & 104 & \textbf{0.9999} & 0.9813 & 353 \\
        \textbf{Mislabel} & 94 & 94 & \textbf{95} & 0.9872 & \textbf{0.3468} & 206 \\
        \hline
    \end{tabular}
    }
    \caption{Random Forgetting 0.15 of data from ResNet.}
    \label{tab:resnet_random_forgetting_with_scrub_0.15_transposed}
\end{table*}

\begin{table*}[h!]
    \centering
    \resizebox{\linewidth}{!}{
    \begin{tabular}{ccccccc}
        \hline
        \textbf{Metric} & \textbf{\boldmath$Acc_t$} & \textbf{\boldmath$Acc_r$} & \textbf{\boldmath$Acc_f$} & \textbf{ZRF} & \textbf{MIA} & \textbf{Time} \\
        \hline
        \textbf{Baseline} & 100 & 100 & 100 & 0.7937 & 0.7608 & - \\
        \textbf{Retrain} & \textbf{100} & \textbf{100} & 99 & 0.8042 & 0.7354 & 686 \\
        \textbf{SSD} & \textbf{100} & \textbf{100} & 100 & 0.7937 & 0.7608 & 255 \\
        \textbf{Incompetent Teacher} & 96 & 96 & \textbf{94} & 0.9890 & \textbf{0.3869} & 202 \\
        \textbf{SCRUB} & 81 & 81 & 104 & \textbf{0.9990} & 0.9315 & 385 \\
        \textbf{Mislabel} & 98 & 98 & 97 & 0.9781 & 0.3882 & \textbf{199} \\
        \hline
    \end{tabular}
    }
    \caption{Random Forgetting 0.25 of data from ResNet.}
    \label{tab:resnet_random_forgetting_with_scrub_0.25_transposed}
\end{table*}

\begin{table*}[h!]
    \centering
    \resizebox{\linewidth}{!}{
    \begin{tabular}{ccccccc}
        \hline
        \textbf{Metric} & \textbf{\boldmath$Acc_t$} & \textbf{\boldmath$Acc_r$} & \textbf{\boldmath$Acc_f$} & \textbf{ZRF} & \textbf{MIA} & \textbf{Time} \\
        \hline
        \textbf{Baseline} & 100 & 100 & 100 & 0.7936 & 0.7581 & - \\
        \textbf{Retrain} & \textbf{100} & \textbf{100} & 99 & 0.8080 & 0.7234 & 649 \\
        \textbf{SSD} & \textbf{100} & \textbf{100} & 100 & 0.7936 & 0.7581 & 259 \\
        \textbf{Incompetent Teacher} & 92 & 92 & \textbf{91} & 0.9933 & 0.3652 & \textbf{209} \\
        \textbf{SCRUB} & 78 & 78 & 102 & \textbf{0.9988} & 0.8360 & 418 \\
        \textbf{Mislabel} & 91 & 91 & 93 & 0.9956 & \textbf{0.3216} & 216 \\
        \hline
    \end{tabular}
    }
    \caption{Random Forgetting 0.35 of data from ResNet.}
    \vspace{10pt}\label{tab:resnet_random_forgetting_with_scrub_0.35_transposed}
\end{table*}

\begin{table*}[h!]
    \centering
    \resizebox{\linewidth}{!}{
    \begin{tabular}{ccccccc}
        \hline
        \textbf{Metric} & \textbf{\boldmath$Acc_t$} & \textbf{\boldmath$Acc_r$} & \textbf{\boldmath$Acc_f$} & \textbf{ZRF} & \textbf{MIA} & \textbf{Time} \\
        \hline
        \textbf{Baseline} & 100 & 100 & 100 & 0.7932 & 0.7596 & - \\
        \textbf{Retrain} & \textbf{100} & \textbf{100} & 99 & 0.8111 & 0.7188 & 612 \\
        \textbf{SSD} & \textbf{100} & \textbf{100} & 100 & 0.7932 & 0.7596 & 269 \\
        \textbf{Incompetent Teacher} & 90 & 90 & \textbf{88} & 0.9962 & \textbf{0.3660} & 217 \\
        \textbf{SCRUB} & 74 & 74 & 95 & 0.9974 & 0.6923 & 461 \\
        \textbf{Mislabel} & 95 & 95 & 95 & 0.9827 & 0.3759 & \textbf{204} \\
        \hline
    \end{tabular}
    }
    \caption{Random Forgetting 0.45 of data from ResNet.}
    \label{tab:resnet_random_forgetting_with_scrub_0.45_transposed}
\end{table*}

\begin{table*}[h!]
    \centering
    \resizebox{\linewidth}{!}{
    \begin{tabular}{ccccccc}
        \hline
        \textbf{Method} & \textbf{\boldmath$Acc_t$} & \textbf{\boldmath$Acc_r$} & \textbf{\boldmath$Acc_f$} & \textbf{ZRF} & \textbf{MIA} & \textbf{Time} \\
        \hline
        \textbf{Baseline} & 100.00 & 100.00 & 100.00 & 0.8646 & 0.802 & - \\
        \textbf{Retrain} & \textbf{100.26} & \textbf{100.26} & 99.99 & 0.8502 & 0.8524 & 7875 \\
        \textbf{SSD} & 100.00 & 100.00 & 99.83 & 0.8694 & 0.8116 & 1760 \\
        \textbf{Incompetent Teacher} & 89.17 & 89.17 & 80.17 & \textbf{0.9937} & \textbf{0.278} & \textbf{1356} \\
        \textbf{Mislabel} & 94.98 & 94.98 & \textbf{86.14} & 0.9347 & 0.3772 & 4032 \\
        \hline
    \end{tabular}
    }
    \caption{Random forgetting results on ViT model (0.05).}
    \label{vit_random_0.05_transposed}
\end{table*}

\begin{table*}[h!]
    \centering
    \resizebox{\linewidth}{!}{
    \begin{tabular}{ccccccc}
        \hline
        \textbf{Method} & \textbf{\boldmath$Acc_t$} & \textbf{\boldmath$Acc_r$} & \textbf{\boldmath$Acc_f$} & \textbf{ZRF} & \textbf{MIA} & \textbf{Time} \\
        \hline
        \textbf{Baseline} & 100.00 & 100.00 & 100.00 & 0.8645 & 0.8109 & - \\
        \textbf{Retrain} & \textbf{100.25} & \textbf{100.25} & \textbf{100.26} & 0.8531 & 0.842 & 7296 \\
        \textbf{SSD} & 100.01 & 100.01 & 100.01 & 0.8694 & 0.8049 & 1939 \\
        \textbf{Incompetent Teacher} & 89.17 & 89.17 & 87.54 & \textbf{0.996} & 0.3132 & \textbf{1553} \\
        \textbf{Mislabel} & 94.13 & 94.13 & 87.93 & 0.9574 & \textbf{0.3827} & 4122 \\
        \hline
    \end{tabular}
    }
    \caption{Random forgetting results on ViT model (0.15).}
    \label{vit_random_0.15_transposed}
\end{table*}

\begin{table*}[h!]
    \centering
    \resizebox{\linewidth}{!}{
    \begin{tabular}{ccccccc}
        \hline
        \textbf{Method} & \textbf{\boldmath$Acc_t$} & \textbf{\boldmath$Acc_r$} & \textbf{\boldmath$Acc_f$} & \textbf{ZRF} & \textbf{MIA} & \textbf{Time} \\
        \hline
        \textbf{Baseline} & 100.00 & 100.00 & 98.74 & 0.8645 & 0.8102 & - \\
        \textbf{Retrain} & \textbf{100.19} & \textbf{100.19} & \textbf{100.12} & 0.8569 & 0.8375 & 6735 \\
        \textbf{SSD} & 100.01 & 100.01 & 99.91 & 0.8697 & 0.8047 & 2105 \\
        \textbf{Incompetent Teacher} & 88.92 & 88.92 & \textbf{1.01} & \textbf{86.40} & 0.3158 & \textbf{1736} \\
        \textbf{Mislabel} & 95.15 & 95.15 & 88.99 & 0.9696 & \textbf{0.3555} & 4240 \\
        \hline
    \end{tabular}
    }
    \caption{Random forgetting results on ViT model (0.25).}
    \label{vit_random_0.25_transposed}
        \vspace{10pt}

\end{table*}

\begin{table*}[h!]
    \centering
    \resizebox{\linewidth}{!}{
    \begin{tabular}{ccccccc}
        \hline
        \textbf{Method} & \textbf{\boldmath$Acc_t$} & \textbf{\boldmath$Acc_r$} & \textbf{\boldmath$Acc_f$} & \textbf{ZRF} & \textbf{MIA} & \textbf{Time} \\
        \hline
        \textbf{Baseline} & 100.00 & 100.00 & 100.00 & 0.8646 & 0.8088 & - \\
        \textbf{Retrain} & \textbf{100.20} & \textbf{100.20} & \textbf{100.10} & 0.8604 & 0.8304 & 6136 \\
        \textbf{SSD} & 100.01 & 100.01 & 99.92 & 0.8695 & 0.8049 & 2278 \\
        \textbf{Incompetent Teacher} & 88.12 & 88.12 & 87.23 & \textbf{0.9978} & 0.3351 & \textbf{1929} \\
        \textbf{Mislabel} & 94.20 & 94.20 & 88.82 & 0.9851 & \textbf{0.2940} & 4384 \\
        \hline
    \end{tabular}
    }
    \caption{Random forgetting results on ViT model (0.35).}
    \label{vit_random_0.35_transposed}
\end{table*}

\begin{table*}[h!]
    \centering
    \resizebox{\linewidth}{!}{
    \begin{tabular}{ccccccc}
        \hline
        \textbf{Method} & \textbf{\boldmath$Acc_t$} & \textbf{\boldmath$Acc_r$} & \textbf{\boldmath$Acc_f$} & \textbf{ZRF} & \textbf{MIA} & \textbf{Time} \\
        \hline
        \textbf{Baseline} & 100.00 & 100.00 & 100.00 & 0.8645 & 0.8103 & - \\
        \textbf{Retrain} & \textbf{100.21} & \textbf{100.21} & \textbf{100.08} & 0.8646 & 0.8222 & 5572 \\
        \textbf{SSD} & 100.01 & 100.01 & 99.91 & 0.8696 & 0.8058 & 2443 \\
        \textbf{Incompetent Teacher} & 85.84 & 85.84 & 85.26 & \textbf{0.9988} & 0.3366 & \textbf{2108} \\
        \textbf{Mislabel} & 94.20 & 94.20 & 88.80 & 0.9851 & \textbf{0.2940} & 4406 \\
        \hline
    \end{tabular}
    }
    \caption{Random forgetting results on ViT model (0.45).}
    \label{vit_random_0.45_transposed}
\end{table*}

\subsubsection{ResNet18 Results}
Most unlearning algorithms struggle with random forgetting, regardless of forget set size. SSD, however, maintains consistent test, retain, and forget accuracy, unlike Mislabel, Incompetent Teacher, and SCRUB, which fail to preserve model performance. SCRUB leads to a significant accuracy drop despite increased forget accuracy, due to issues with distinguishing forget and retain sets. Selective Synaptic Dampening fails to forget across all percentage values, likely due to similarities in Fisher Information Matrix values, while the naive retrain shows only minimal changes in MIA and ZRF metrics.

\subsubsection{ViT Results}
Similar to the results of ResNet18, the random forgetting experiment shows that both the naive retrain and Selective Synaptic Dampening fail to achieve unlearning at any percentage value. In contrast, Incompetent Teacher and Mislabel successfully achieve unlearning across all percentages. Incompetent Teacher performs better at smaller percentages but retains less model accuracy than Mislabel. Mislabel, however, excels at higher percentages, offering better unlearning performance and model accuracy retention, though at the cost of more than twice the time required for the unlearning procedure.\
Second, text classification models.
\subsection{MARBERT Results}
In the Full-Class Forgetting
experiments, MARBERT was fine-tuned on
HARD dataset. Since classes vary significantly in size, unlearning experiments were done for each of the 5 classes in HARD.
We note that Selective Synaptive Dampening achieves the highest retain accuracy values with very strong MIA and ZRF values for classes 1, 2, and 3 but more time taken compared to other methods and fails to unlearn in both classes 4, and 5 corresponding to the largest portion of the dataset. 
Evaluating Incompetent Teacher, we note that for classes 1 and 4 we obtain higher accuracy retention with near equivalent unlearning performance by utilizing a smaller learning rate, whereas for class 2, the higher learning rate provides better accuracy retention with a minuscule decrease in unlearning performance and for classes 3 and 5, performance is equivalent between both learning rates. Overall, incompetent teacher manages to achieve unlearning but model performance starts to degrade sharply for classes 4 and 5 which contain large portions of the training dataset.

Evaluating Mislabel Unlearning, we note very high degradation in model performance across all five classes, but achieves complete forgetting at the cost of model prediction randomness.

Evaluating UNSIR, we note catastrophic performance by the algorithm using 1e-4 and 2e-4 learning rates respectively in forgetting class 1, and catastrophic performance by the higher learning rate in forgetting class 2 as well, with the MIA value rising to 1.0. The MIA value for the higher learning rate inflects to 0 for classes 3, 4, and 5 however with near Mislabel Unlearning accuracy retention. For class 2, the lower learning rate achieves slightly worse than Incompetent Teacher accuracy retention with good MIA and ZRF metrics at a lower time taken, and for class 3 accuracy retention remains slightly below Incompetent Teacher but MIA and ZRF values degrade. For classes 4 and 5 the lower learning rate option fails to achieve unlearning.
SCRUB, despite being top performer in Full-class forgetting in ResNet, and one of the best methods with ViT, failed to adapt to new task and achived worst results among all algorithms, overall model accuracy dropped to less than half, highlighting algorithm's inability to unlearn textual information.
We note as well the failure of the naive retrain to achieve unlearning in all but class 4 where it achieves the best retain accuracy but with slightly higher MIA values and lower ZRF values.

Overall, for this experiment we identify a failure by most algorithms to provide satisfactory unlearning, with Incompetent Teacher providing the best all around results using differing learning rates.  We attribute the difficulty in achieving unlearning during this experiment to the imbalanced nature of the dataset used.

\begin{table*}[h!]
    \centering
    \resizebox{\linewidth}{!}{
    \begin{tabular}{cccccccccc}
        \hline
        \textbf{Metric} & \textbf{\boldmath$Acc_t$} & \textbf{\boldmath$Acc_r$} & \textbf{\boldmath$Acc_f$} & \textbf{ZRF} & \textbf{MIA} & \textbf{Time} \\
        \hline
        \textbf{Baseline} & 100.00 & 100.00 & 100.00 & 1.00 & 0.4046 & - \\
        \textbf{Retrain} & \textbf{99.39} & \textbf{100.01} & 66.45 & 0.9894 & 0.5523 & 5603 \\
        \textbf{SSD} & 97.93 & 99.73 & 2.90 & 0.9214 & 0.0195 & 3412 \\
        \textbf{Incompetent Teacher} & 85.82 & 87.44 & \textbf{0} & 0.9547 & 0.2351 & 2238 \\
        \textbf{SCRUB} & 43.00 & 42.01 & \textbf{0} & \textbf{1.00} & 0.6538 & 7191 \\
        \textbf{UNSIR} & 84.34 & 86.00 & \textbf{0} & 0.6834 & 0.2132 & \textbf{1909} \\
        \textbf{Mislabel} & 41.54 & 42.35 & \textbf{0} & 0.6685 & \textbf{0} & 3568 \\
        \hline
    \end{tabular}
    }
    \caption{Results for Forgetting Class 1 of HARD dataset from MARBERT}
    \label{tab:MARBERT class 1 transposed rounded}
\end{table*}

\begin{table*}[h!]
    \centering
    \resizebox{\linewidth}{!}{
    \begin{tabular}{cccccccccc}
        \hline
        \textbf{Metric} & \textbf{\boldmath$Acc_t$} & \textbf{\boldmath$Acc_r$} & \textbf{\boldmath$Acc_f$} & \textbf{ZRF} & \textbf{MIA} & \textbf{Time} \\
        \hline
        \textbf{Baseline} & 100.00 & 100.00 & 100.00 & 1.00 & 0.4617 & - \\
        \textbf{Retrain} & \textbf{97.17} & \textbf{102.42} & 45.88 & 0.9612 & 0.4777 & 5315 \\
        \textbf{SSD} & 86.97 & 95.91 & \textbf{0} & 0.8389 & 0.0188 & 3597 \\
        \textbf{Incompetent Teacher} & 85.60 & 94.17 & 2.35 & 0.9875 & 0.0197 & 2392 \\
        \textbf{SCRUB} & 41.56 & 45.16 & \textbf{0} & \textbf{1.00} & 0.4448 & 7250 \\
        \textbf{UNSIR} & 75.90 & 83.82 & \textbf{0} & 0.6716 & 0.2112 & \textbf{1911} \\
        \textbf{Mislabel} & 41.54 & 45.85 & \textbf{0} & 0.62 & \textbf{0} & 3584 \\
        \hline
    \end{tabular}
    }
    \caption{Results for Forgetting Class 2 of HARD dataset from MARBERT}
    \label{tab:MARBERT class 2 transposed rounded}
        \vspace{10pt}

\end{table*}

\begin{table*}[h!]
    \centering
    \resizebox{\linewidth}{!}{
    \begin{tabular}{cccccccccc}
        \hline
        \textbf{Metric} & \textbf{\boldmath$Acc_t$} & \textbf{\boldmath$Acc_r$} & \textbf{\boldmath$Acc_f$} & \textbf{ZRF} & \textbf{MIA} & \textbf{Time} \\
        \hline
        \textbf{Baseline} & 100.00 & 100.00 & 100.00 & 1.00 & 0.5408 & - \\
        \textbf{Retrain} & 92.21 & \textbf{104.47} & 45.60 & 0.9532 & 0.4109 & 4877 \\
        \textbf{SSD} & \textbf{93.03} & 104.86 & 48.02 & 0.9441 & 0.0891 & 3838 \\
        \textbf{Incompetent Teacher} & 75.03 & 94.62 & \textbf{0} & 0.9949 & 0.0408 & 2601 \\
        \textbf{SCRUB} & 42.32 & 43.50 & \textbf{0} & \textbf{1.00} & 0.4452 & 7206 \\
        \textbf{UNSIR} & 78.91 & 99.64 & \textbf{0} & 0.6938 & 0.4692 & \textbf{1911} \\
        \textbf{Mislabel} & 41.54 & 52.52 & \textbf{0} & 0.6165 & \textbf{0} & 3601 \\
        \hline
    \end{tabular}
    }
    \caption{Results for Forgetting Class 3 of HARD dataset from MARBERT}
    \label{tab:MARBERT class 3 transposed rounded}
\end{table*}

\begin{table*}[h!]
    \centering
    \resizebox{\linewidth}{!}{
    \begin{tabular}{cccccccccc}
        \hline
        \textbf{Metric} & \textbf{\boldmath$Acc_t$} & \textbf{\boldmath$Acc_r$} & \textbf{\boldmath$Acc_f$} & \textbf{ZRF} & \textbf{MIA} & \textbf{Time} \\
        \hline
        \textbf{Baseline} & 100.00 & 100.00 & 100.00 & 1.00 & 0.5174 & - \\
        \textbf{Retrain} & 75.82 & \textbf{111.66} & 1.55 & 0.8891 & 0.2407 & 4492 \\
        \textbf{SSD} & \textbf{100.32} & 100.19 & 100.59 & 0.9998 & 0.5162 & 4065 \\
        \textbf{Incompetent Teacher} & 69.49 & 78.84 & 50.60 & 0.9942 & 0.0076 & 2828 \\
        \textbf{SCRUB} & 42.67 & 42.80 & \textbf{0} & 0.6685 & 0.4449 & 7269 \\
        \textbf{UNSIR} & 62.97 & 93.50 & \textbf{0} & 0.6822 & 0.4397 & \textbf{1913} \\
        \textbf{Mislabel} & 43.09 & 63.80 & \textbf{0} & 0.6427 & \textbf{0} & 3618 \\
        \hline
    \end{tabular}
    }
    \caption{Results for Forgetting Class 4 of HARD dataset from MARBERT}
    \label{tab:MARBERT class 4 transposed rounded}
\end{table*}

\begin{table*}[h!]
    \centering
    \resizebox{\linewidth}{!}{
    \begin{tabular}{cccccccccc}
        \hline
        \textbf{Metric} & \textbf{\boldmath$Acc_t$} & \textbf{\boldmath$Acc_r$} & \textbf{\boldmath$Acc_f$} & \textbf{ZRF} & \textbf{MIA}

 & \textbf{Time} \\
        \hline
        \textbf{Baseline} & 100.00 & 100.00 & 100.00 & 1.00 & 0.6694 & - \\
        \textbf{Retrain} & 88.13 & \textbf{103.86} & 35.40 & 0.8916 & 0.5412 & 4963 \\
        \textbf{SSD} & 82.54 & 100.91 & 38.69 & 0.9456 & 0.0154 & 3845 \\
        \textbf{Incompetent Teacher} & 62.77 & 96.98 & \textbf{0} & 0.9167 & 0.4101 & 2685 \\
        \textbf{SCRUB} & \textbf{43.55} & 43.82 & \textbf{0} & \textbf{1.00} & 0.4617 & 7325 \\
        \textbf{UNSIR} & 78.65 & 94.71 & \textbf{0} & 0.6338 & 0.5612 & \textbf{1927} \\
        \textbf{Mislabel} & 44.32 & 47.89 & \textbf{0} & 0.6695 & \textbf{0} & 3611 \\
        \hline
    \end{tabular}
    }
    \caption{Results for Forgetting Class 5 of HARD dataset from MARBERT}
    \label{tab:MARBERT class 5 transposed rounded}
    \vspace{10pt}
\end{table*}
\section{Conclusion}
Combining the observations and conclusions from the previous experiments, we identify that Incompetent Teacher achieves the most balanced results in all work loads, however
for specific workloads, other methods achieve better results.
Mislabel Unlearning achieves better unlearning performance in full-class forgetting on models utilizing transfer learning, and goes head to head with Incompetent Teacher in
random forgetting work loads.
SCRUB demonstrates superior results in the Full-Class forgetting task on ResNet, and achieves competitive results with ViT. However, SCRUB's adaptation to new tasks, such as Text Classification, proved inadequate, as it yielded low overall model accuracy and retention accuracy. Additionally, SCRUB resulted in catastrophic unlearning when applied to Random Forgetting scenario.
Selective Synaptive Dampening achieves better results in the full-class forgetting process on a model trained from scratch,While UNSIR provides good performance in many of the previous experiments, it lags in accuracy retention and unlearning performance as well as time taken to be an alternative to Incompetent Teacher in any workload. 

\bibliographystyle{plain}
\bibliography{samplebib}

\begin{thebibliography}{10}

\bibitem{abdul-mageed-etal-2021-arbert}
Muhammad Abdul-Mageed, AbdelRahim Elmadany, and El~Moatez~Billah Nagoudi.
\newblock {ARBERT} {\&} {MARBERT}: Deep bidirectional transformers for {A}rabic.
\newblock In {\em Proceedings of the 59th Annual Meeting of the Association for Computational Linguistics and the 11th International Joint Conference on Natural Language Processing (Volume 1: Long Papers)}, pages 7088--7105, Online, August 2021. Association for Computational Linguistics.

\bibitem{chundawat2023zeroshot}
Vikram~S Chundawat, Ayush~K Tarun, Murari Mandal, and Mohan Kankanhalli.
\newblock Zero-shot machine unlearning, 2023.

\bibitem{dosovitskiy2021image}
Alexey Dosovitskiy, Lucas Beyer, Alexander Kolesnikov, Dirk Weissenborn, Xiaohua Zhai, Thomas Unterthiner, Mostafa Dehghani, Matthias Minderer, Georg Heigold, Sylvain Gelly, Jakob Uszkoreit, and Neil Houlsby.
\newblock An image is worth 16x16 words: Transformers for image recognition at scale, 2021.

\bibitem{Elnagar2018}
Ashraf Elnagar, Yasmin~S. Khalifa, and Anas Einea.
\newblock {\em Hotel Arabic-Reviews Dataset Construction for Sentiment Analysis Applications}, pages 35--52.
\newblock Springer International Publishing, Cham, 2018.

\bibitem{foster2023fast}
Jack Foster, Stefan Schoepf, and Alexandra Brintrup.
\newblock Fast machine unlearning without retraining through selective synaptic dampening, 2023.

\bibitem{graves2021amnesiac}
Laura Graves, Vineel Nagisetty, and Vijay Ganesh.
\newblock Amnesiac machine learning.
\newblock In {\em Proceedings of the AAAI Conference on Artificial Intelligence}, volume~35, pages 11516--11524, 2021.

\bibitem{he2016deep}
Kaiming He, Xiangyu Zhang, Shaoqing Ren, and Jian Sun.
\newblock Deep residual learning for image recognition.
\newblock In {\em Proceedings of the IEEE conference on computer vision and pattern recognition}, pages 770--778, 2016.

\bibitem{krizhevsky2009learning}
Alex Krizhevsky, Geoffrey Hinton, et~al.
\newblock Learning multiple layers of features from tiny images.
\newblock 2009.

\bibitem{kurmanji2023unbounded}
Meghdad Kurmanji, Peter Triantafillou, Jamie Hayes, and Eleni Triantafillou.
\newblock Towards unbounded machine unlearning, 2023.

\bibitem{lin1991divergence}
J.~Lin.
\newblock Divergence measures based on the shannon entropy.
\newblock {\em IEEE Transactions on Information Theory}, 37(1):145--151, 1991.

\bibitem{shokri2017membership}
Reza Shokri, Marco Stronati, Congzheng Song, and Vitaly Shmatikov.
\newblock Membership inference attacks against machine learning models, 2017.

\bibitem{tarun2023fast}
Ayush~K. Tarun, Vikram~S. Chundawat, Murari Mandal, and Mohan Kankanhalli.
\newblock Fast yet effective machine unlearning.
\newblock {\em IEEE Transactions on Neural Networks and Learning Systems}, page 1–10, 2023.

\bibitem{wu2020visual}
Bichen Wu, Chenfeng Xu, Xiaoliang Dai, Alvin Wan, Peizhao Zhang, Zhicheng Yan, Masayoshi Tomizuka, Joseph Gonzalez, Kurt Keutzer, and Peter Vajda.
\newblock Visual transformers: Token-based image representation and processing for computer vision, 2020.

\end{thebibliography}
\end{document}